\documentclass[final]{cvpr}

\usepackage{times}
\usepackage{epsfig}
\usepackage{graphicx}
\graphicspath{{assets/}} %Setting the graphicspath
\usepackage{amsmath}
\usepackage{amssymb}

\usepackage{fancyref}
\usepackage{cite}
\DeclareMathAlphabet\mathcalbf{OMS}{cmsy}{b}{n}
\usepackage{mathabx}
\usepackage{mathtools}
\usepackage[utf8]{inputenc}
\usepackage[T1]{fontenc}
\usepackage{float}
\DeclareMathOperator*{\argmax}{arg\,max}

\usepackage{appendix}

\usepackage[pagebackref=true,breaklinks=true,colorlinks,bookmarks=false]{hyperref}

\def\cvprPaperID{****} % *** Enter the CVPR Paper ID here
\def\confYear{CVPR 2021}

\begin{document}

%%%%%%%%% TITLE
\title{Combining Semantic Guidance and Deep Reinforcement Learning For Generating Human Level Paintings}

\author{Jaskirat Singh\\
Australian National University\\
% Canberra, Australia\\
{\tt\small jaskirat.singh@anu.edu.au}
\and
Liang Zheng\\
Australian National University\\
% Canberra, Australia\\
{\tt\small liang.zheng@anu.edu.au}
}

% {\small\begin{verbatim}
%   \usepackage[dvips]{graphicx} ...
%   \includegraphics[width=0.8\linewidth]
%                   {myfile.eps}
% \end{verbatim}
% }

\maketitle

%------------------------------------------------------------------%
%%%%%%%%% ABSTRACT
\begin{abstract}
    Generation of stroke-based non-photorealistic imagery, is an important problem in the computer vision community. As an endeavor in this direction, substantial recent research efforts have been focused on teaching machines ``how to paint'', in a manner similar to a human painter. However, the applicability of previous methods has been limited to datasets with little variation in position, scale and saliency of the foreground object. As a consequence, we find that these methods struggle to cover the granularity and diversity possessed by real world images. To this end, we propose a Semantic Guidance pipeline with 1) a bi-level painting procedure for learning the distinction between foreground and background brush strokes at training time. 2) We also introduce invariance to the position and scale of the foreground object through a neural alignment model, which combines object localization and spatial transformer networks in an end to end manner, to zoom into a particular semantic instance. 3) The distinguishing features of the in-focus object are then amplified by maximizing a novel guided backpropagation based focus reward. The proposed agent does not require any supervision on human stroke-data and successfully handles variations in foreground object attributes, thus, producing much higher quality canvases for the CUB-200 Birds \cite{WahCUB_200_2011} and Stanford Cars-196 \cite{KrauseStarkDengFei-Fei_3DRR2013} datasets. Finally, we demonstrate the further efficacy of our method on complex datasets with multiple foreground object instances by evaluating an extension of our method on the challenging Virtual-KITTI \cite{cabon2020vkitti2} dataset. Source code and models are available at \url{https://github.com/1jsingh/semantic-guidance}.
\end{abstract}

%------------------------------------------------------------------%
%%%%%%%%% BODY TEXT
\section{Introduction}
\label{sec:introduction}

\begin{figure}[h]
% \vskip 0.2in
\begin{center}
\centerline{\includegraphics[width=0.98\linewidth]{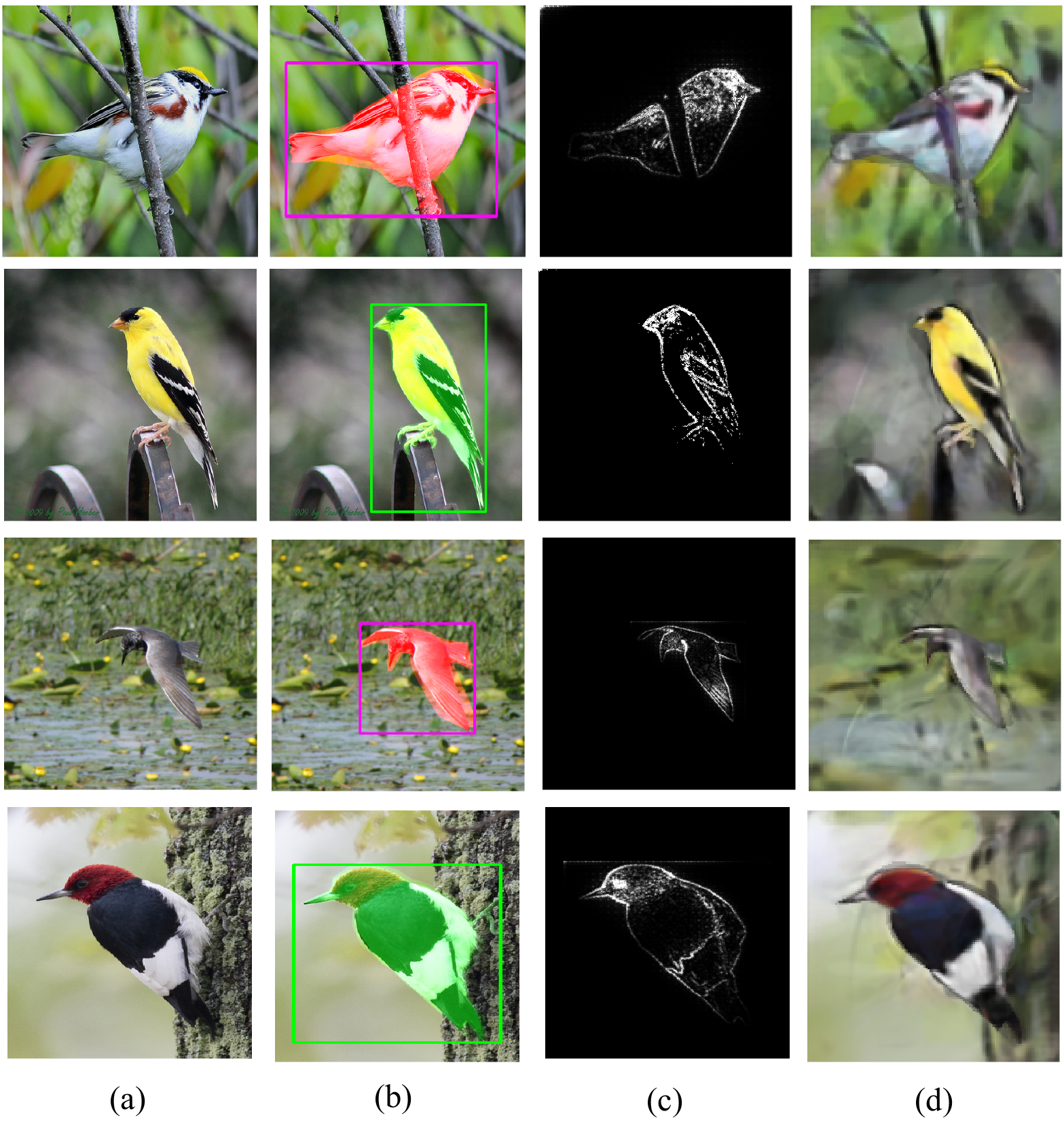}}
\caption{\textbf{Semantic Guidance}. We propose a semantic guidance pipeline for the \emph{``learning to paint''} problem. The reinforcement learning agent incorporates (b) object localization and semantic segmentation maps for the target image (a), to achieve enhanced foreground saliency (refer Fig.~\ref{fig:bird_results}) in the final canvas (d). We also introduce expert guidance to amplify the focus on small but distinguishing features of the foreground objects (\emph{e.g.} bird's eye), by proposing (c) a guided backpropagation based focus reward.}
\label{fig:overview}
\end{center}
\vskip -0.3in
\end{figure}

Paintings form a key medium through which humans express their visual conception, creativity and thoughts. Being able to paint constitutes a vital skill in the human learning process and requires long-term planning to efficiently convey the picture within a limited number of brush strokes. Thus, the successful impartation of this challenging skill to machines, would not only have huge applications in computer graphics, but would also form a key component in the development of a general artificial intelligence system.

Recently, a lot of research \cite{huang2019learning,mellor2019unsupervised,ganin2018synthesizing,zheng2018strokenet,xie2013artist,ha2017neural} is being targeted on teaching machines \emph{``how to paint''}, in a manner similar to a human painter. A popular solution to this problem is to use reinforcement learning and model the painting episode as a Markov Decision Process (MDP).
Given a target image, the agent learns to predict a sequence of brush strokes which when transferred on to a canvas, result in a painting which is semantically and visually similar to the input image. The reward function for the agent is usually learnt using a generative adversarial network (GAN) \cite{goodfellow2014generative}, which provides a measure of similarity between the final canvas and the original target image.

In this paper, we propose a \emph{semantic guidance} pipeline which addresses the following three challenges faced by the current painting agents. \textbf{First}, the current methods \cite{huang2019learning,mellor2019unsupervised,ganin2018synthesizing} are limited to only datasets which depict a single dominant instance per image (\emph{e.g.}~cropped faces). Experimental results reveal that this leads to poor performance on varying the position, scale and saliency of the foreground object within the image. We address this limitation by adopting a \emph{bi-level painting procedure}, which incorporates semantic segmentation into the painting process, to learn a distinction between brush stroke patterns for foreground and background image regions. Here, we utilize the intuition that the human painting process is deeply rooted in our semantic understanding of the image components. For instance, an accurate depiction of a bird sitting on a tree would depend highly on the agent's ability to recognize the bird and the tree as separate objects and hence use correspondingly different stroke patterns / plans.

\textbf{Second}, variation in position and scale of the foreground objects within the image, introduces high variance in the input distribution for the generative model. To this end, we propose a \emph{neural alignment model}, which combines object localization and spatial transformer networks to learn an affine mapping between the overall image and the bounding box of the target object. The neural alignment model is end-to-end and preserves the differentiability requirement for our model-based reinforcement learning approach.

\textbf{Third}, accurate depiction of instances belonging to the same semantic class should require the painting agent to give special attention to different distinguishing features. For instance, while the shape of the beak may be a key feature for some birds, it may be of little consequence for other bird types. We thus propose a novel guided backpropagation based \emph{focus reward} to increase the model's attention on these fine-grain features. The use of guided backpropagation also helps in amplifying the importance of small image regions, like a bird's eye which might be otherwise ignored by the reinforcement learning agent.

In summary, the main contributions of this paper are:
\begin{itemize}
    \item We introduce a semantically guided bi-level painting process to develop a better distinction between foreground and background brush stroke patterns.
    \item We propose a neural alignment model, which combines object localization and spatial transformer networks in an end to end manner to zoom in on a particular foreground object in the image.
    \item We finally introduce expert guidance on the relative importance of distinguishing features of the in-focus object (\emph{e.g.} tail, beak \emph{etc.}~for a bird) by proposing a novel guided backpropagation based focus reward.
\end{itemize}

%------------------------------------------------------------------%
% \newpage
\section{Related Work}
\label{sec:related_work}

\textbf{Stroke based rendering methods.} Automatic generation of non-photorealistic imagery has been a problem of keen interest in the computer vision community. Stroke Based Rendering (SBR) is a popular approach in this regard, which focuses on recreating images by placing discrete elements such as paint strokes or stipples \cite{hertzmann2003survey}.

The positioning and selection of appropriate strokes is a key aspect of this approach \cite{zeng2009image}.
Most traditional SBR algorithms address this task through either, greedy search at each step \cite{hertzmann1998painterly,litwinowicz1997processing}, optimization over an energy function using heuristics \cite{turk1996image}, or require user interaction for supervising brush stroke positions \cite{haeberli1990paint,teece19983d}.

\textbf{RNN-based methods.} Recent deep learning based solutions adopt the use of recurrent neural networks for stroke decomposition. However, these methods like Sketch-RNN \cite{ha2017neural} for drawings and Graves \etal \cite{graves2013generating} for handwriting generation, require access to sequential stroke data, which limits their applicability for most real world datasets. StrokeNet \cite{ha2017neural} addresses this limitation by using a differentiable renderer, however it fails to generalize to color images.

\textbf{Unsupervised stroke decomposition using RL.} Recent methods \cite{xie2013artist,ganin2018synthesizing,mellor2019unsupervised,huang2019learning} use RL to learn an efficient stroke decomposition. The adoption of a trial and error approach alleviates the need for stroke supervision, as long as a reliable reward metric is available. SPIRAL \cite{ganin2018synthesizing}, SPIRAL++ \cite{mellor2019unsupervised} and Huang \etal \cite{huang2019learning} adopt an adversarial training approach, wherein the reward function is modelled using the WGAN distance \cite{huang2019learning,arjovsky2017wasserstein}. 
% However, a pure RL based approach.
Learning a differentiable renderer model has also been shown to improve the learning speed of the training process \cite{huang2019learning,zheng2018strokenet,nakano2019neural,frans2018unsupervised}. 

The above methods generalize only for datasets (\emph{e.g.} cropped, aligned faces from CelebA \cite{liu2015faceattributes}), with limited variation in scale, position and saliency of the foreground object. We note that while Huang \etal \cite{huang2019learning} evaluate their approach on ImageNet \cite{deng2009imagenet}, we find that competitive results are achieved only after using the division parameter at inference times. In this setting, the agent divides the overall image into a grid with 16 / 256 blocks, and then proceeds to paint each of them in parallel. We argue that such a division does not follow the constraints of the original problem formulation, in which the agent mimics the human painting process. Furthermore, such a division strategy increases the effective number of total strokes and tends towards a pixel-level image regression approach, with the generated images losing the desired artistic / non-photorealistic touch.

\textbf{Semantic Divide and Conquer.}
Our work is in part also motivated by semantic division strategies from \cite{wang2020sdc,liu2010single}, which propose a division of the overall depth estimation task among the constituent semantic classes. However, to the best of our knowledge, our work is the first attempt on incorporating semantic division (with model-based RL) for the ``learning to paint'' problem.

\begin{figure*}[h!]
% \vskip 0.2in
\begin{center}
% \centerline{\includegraphics[width=0.98\linewidth,height=4cm]{bird_results1.pdf}}
\centerline{\includegraphics[width=0.9\linewidth]{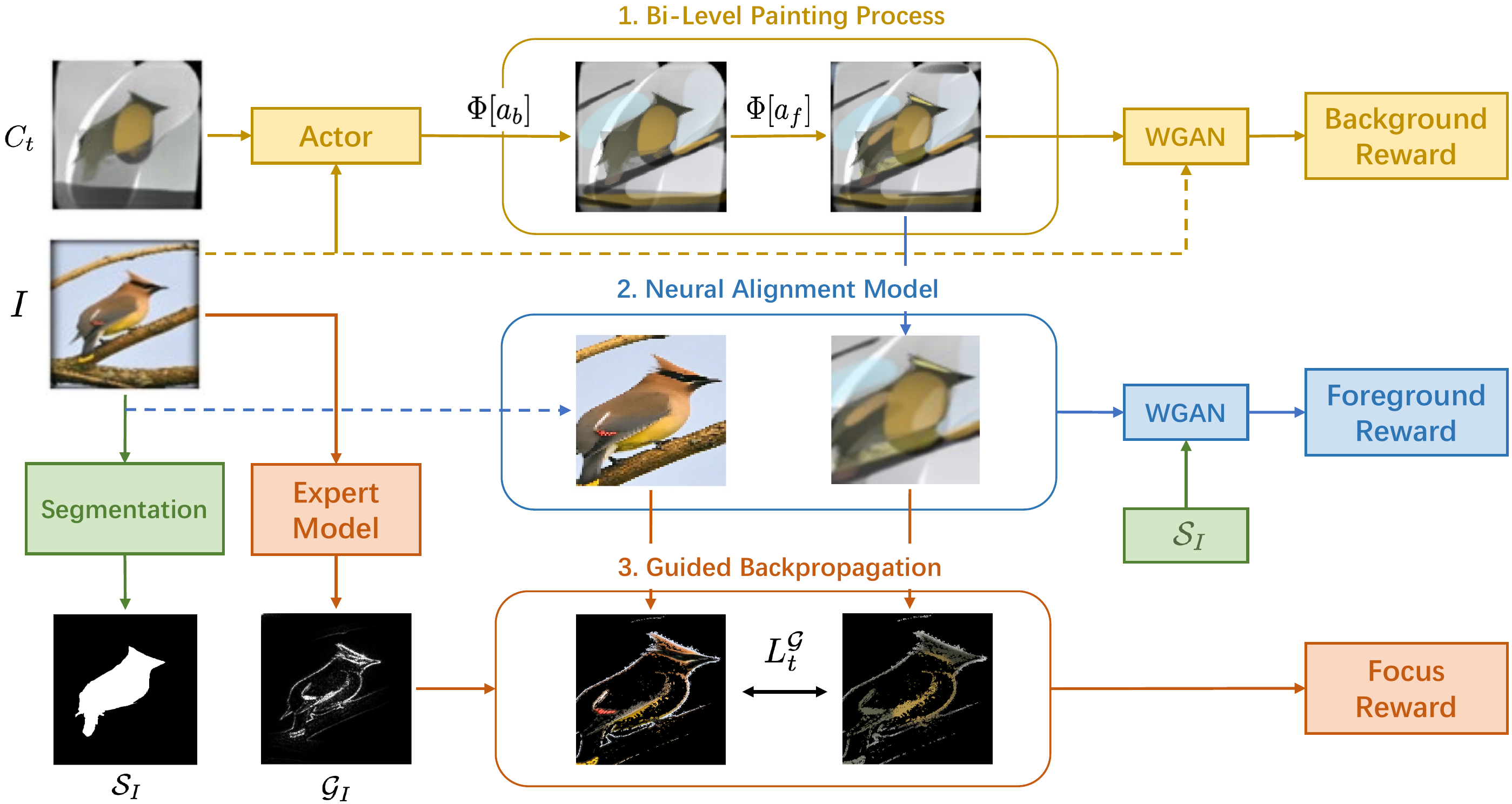}}
\caption{\textbf{Overview of Semantic Guidance Pipeline.} Our semantic guidance pipeline consists of three parts. \textbf{1)} The bi-level painting process (Section \ref{sec:bilevel}) develops a distinction between painting foreground and background brush strokes. \textbf{2)} The Neural Alignment Model (Section \ref{sec:neural_alignment}) provides a differentiable cropping of the foreground object regions for the target image and the updated canvas state. These cropped object images are then used to compute the foreground reward (refer Eq. \ref{eq:fg_reward2}). \textbf{3)} Finally, we use guided backpropagation maps from an expert model, to specifically boost the importance of distinguishing object features in the final canvas (Section \ref{sec:focus_reward}).}
\label{fig:model_design}
\end{center}
\vskip -0.3in
\end{figure*}

%------------------------------------------------------------------%
\section{Overview of the Painting Agent}
\label{sec:overview}

% \subsection{Basic Setup}
Similar to Huang \etal \cite{huang2019learning}, we adopt a model-based reinforcement learning approach for this problem. The painting episode is modelled as a Markov Decision Process (MDP) defined by state space $\mathcal{S}$, transition function $\mathcal{P}(s_{t+1}|s_t,a_t)$ and action space $\mathcal{A}$. 

\textbf{State space.} The state $s_t \in \mathcal{S}$ at any time $t$ is defined by the tuple $(C_t,I,\mathcalbf{S}_I,\mathcalbf{G}_I,t)$, where $C_t$ is the canvas image at timestep $t$ and $I$ is the target image. $\mathcalbf{S}_I,\mathcalbf{G}_I$ represent the semantic instance probability map $\{\in [0,1]^{H \times W}\}$ and the guided backpropagation map for the target image.
% $\{\in \{0,1\}^{H \times W}\}$

\textbf{Action space}. The action $a_t$ at each timestep, depicts the parameters of a quadratic Bézier curve, used to model the brush stroke. The stroke parameters form a 13 dimensional vector as follows, 
\begin{align}
    a_t = (x_0,y_0,x_1,y_1,x_2,y_2,z_0,z_2,w_0,w_2,r,g,b),
\end{align}
where the first 10 parameters depict stroke position, shape and transparency, while the last 3 parameters $(r,g,b)$ form the RGB representation for the stroke color.

\textbf{Environment Model.} The environment model / transition function $\mathcal{P}(s_{t+1}|s_t,a_t)$ is modelled through a neural renderer network $\mathbf{\Phi}$, which facilitates a differentiable mapping from the current canvas $C_t$ and brush stroke parameters $a_t$ to the updated canvas state $C_{t+1}$. For mathematical convenience alone, we define two distinct stroke map definitions $\mathbf{\Phi},\mathbf{\Phi}^c$ . $\mathbf{\Phi}(a_t) \{ \in [0,1]^{H \times W}\}$ represents the stroke density map, whose value at any pixel provides a measure of transparency of the current stroke. $\mathbf{\Phi}^c(a_t)$ is the colored rendering of the original stroke density map $\mathbf{\Phi}(a_t)$ on an empty canvas.

\textbf{Action Bundle.} We adopt an action bundle approach which has been shown to be an efficient mechanism for enforcing higher emphasis on the planning process \cite{huang2019learning}. Thus, at each timestep the agent predicts the parameters for the next $K=5$ brush strokes.

\section{Introducing Semantic Guidance}
\label{sec:semantic_guidance}

In the following sections, we describe the complete pipeline for our semantic guidance model (refer Fig.~\ref{fig:model_design}). We first outline our approach for a two class (foreground, background) painting problem and then later demonstrate its extension to more complex image datasets with multiple foreground instances per image in Section \ref{sec:multiple_foreground}.

\subsection{The Bi-Level Painting Process}
\label{sec:bilevel}
% \subsection{}
The human painting process is inherently multi-level, wherein the painter would focus on different semantic regions through distinct brush strokes. For instance, brush strokes aimed at painting the general image background would have a different distribution as compared to strokes depicting each of the foreground instances. 

Motivated by this, we propose to use semantic segmentation to develop a distinction between the foreground and the background strokes. This distinction is achieved through a bi-level painting procedure which allocates a specialized reward for each stroke type. More specifically, we first modify the action bundle $\mathbf{a}_t$ to separately predict Bézier curve parameters for foreground and background strokes, \emph{i.e.}
\begin{align}
    \mathbf{a}_t = \{\mathbf{a}_b,\mathbf{a}_f\},
\end{align}
where $\mathbf{a}_f,\mathbf{a}_b$ represent the foreground and background stroke parameters, respectively.
Next, given a neural renderer network $\mathbf{\Phi}$, target image $I$ and semantic class probability map $\mathcalbf{S}_I$, the canvas state $C_t$ is updated in the following two stages,
\begin{align}
    &C^b_{t+1} =  [1-\mathbf{\Phi}(\mathbf{a}_b)] \odot C_t +  \mathbf{\Phi}^c(\mathbf{a}_b) \odot [1-\mathcalbf{S}_I],\\
    &C_{t+1} = [1-\mathbf{\Phi}(\mathbf{a}_f)] \odot C^b_{t+1} + \mathbf{\Phi}^c(\mathbf{a}_f) \odot \mathcalbf{S}_I,
    % \\C_{t+1} = C^f_{t+1}
\end{align}
where $\odot$ indicates element-wise multiplication and $\mathbf{\Phi}^c(a)$ is the colored rendering of the stroke density map $\mathbf{\Phi}(a)$.

The reward for each stroke type is then defined as,
\begin{align}
    r^b_t = D^{wgan}(I, C_{t+1}) - D^{wgan}(I,C_{t}),
\end{align}
\begin{equation}
    \begin{aligned}
        r^f_t = \ &D^{wgan}(I \odot \mathcalbf{S}_I , C_{t+1} \odot \mathcalbf{S}_I ) \\ 
            - & D^{wgan}(I \odot \mathcalbf{S}_I,C_{t} \odot \mathcalbf{S}_I),
    \end{aligned} \label{eq:fg_reward1}
\end{equation}
where $r^f_t,r^b_t$ represent the foreground and background rewards, respectively, and $D^{wgan}(I,C_t)$ is the joint conditional discriminator score for image $I$ and canvas $C_t$. 

\subsection{Neural Alignment Model}
\label{sec:neural_alignment}
The accuracy of the foreground rewards computed using Eq. \ref{eq:fg_reward1}, depends highly on the ability of the discriminator to accurately capture the similarity between the foreground regions in target image $I$ and the current canvas state $C_t$. However, the input to the discriminator of the WGAN model would have high variance, if the position and scale of the foreground object varies significantly amongst the input images. This high variance poses a direct challenge to the discriminator's performance while training on complex real world datasets. To this end, we propose a differentiable neural alignment model, which combines object localization and spatial transformer networks \cite{jaderberg2015spatial} to zoom into the foreground object, thereby providing a standardized input for the discriminator.

First, we modify the segmentation model to predict both the foreground object mask $\mathcalbf{S}_I$ and bounding box coordinates $(x_b,y_b,w_b,h_b)$ of the foreground object in the target image. We then use a spatial transformer network $\mathbf{\Omega}$, which uses the predicted bounding box coordinates to compute an affine mapping, from the overall canvas image $C_t$ to the zoomed foreground object image $Z^C_t$. Mathematically,
\begin{align}
    &\mathcalbf{S}_I, (x_b,y_b,w_b,h_b) = \mathbf{\Psi} (I),\\
    &Z^C_t = \mathbf{\Omega}(C_t,(x_b,y_b,w_b,h_b)), \\
    &Z^I = \mathbf{\Omega}(I, (x_b,y_b,w_b,h_b)),\\
    &Z^\mathcalbf{S} = \mathbf{\Omega}(\mathcalbf{S}_I, (x_b,y_b,w_b,h_b)),
\end{align}
where $\mathbf{\Psi}$ represents the foreground segmentation and localization network. The $3 \times 2$ affine matrix for the spatial transformer network $\mathbf{\Omega}$, given bounding box coordinates $(x_b,y_b,w_b,h_b)$ and overall image size $(H,W)$, is defined as,
\begin{align}
    A = 
    \begin{bmatrix}
    W/w_b & 0 & -Wx_b/w_b\\
    0 & H/h_b & -Hy_b/h_b
    \end{bmatrix}^T .
\end{align}

The modified foreground reward ($r^f_t$) is then computed using the WGAN discriminator scores for the zoomed-in target and canvas images, as follows,
\begin{equation}
    \begin{aligned}
        r^f_t = \ &D^{wgan}(Z^I \odot Z^\mathcalbf{S}, Z^C_{t+1} \odot Z^\mathcalbf{S}  ) \\ 
            - & D^{wgan}(Z^I \odot Z^\mathcalbf{S},Z^C_{t} \odot Z^\mathcalbf{S}).
    \end{aligned} \label{eq:fg_reward2}
\end{equation}

\subsection{Guided Backpropagation Based Focus Reward}
\label{sec:focus_reward}
The semantic importance of an image region is not necessarily proportional to the number of pixels covered by the corresponding region. While using WGAN loss provides some degree of abstraction as compared with the direct pixel-wise $l_2$ distance, we observe that a painting agent trained with a WGAN score based reward function, does not pay adequate attention to small but distinguishing object features. For instance, as shown in Fig.~\ref{fig:bird_results}, for the CUB-200-2011 birds dataset, we see that while the baseline agent captures the global object features like shape and color, it either omits or insufficiently depicts important bird features like eyes, wing texture, color marks around the neck \emph{etc}.

In order to address this limitation, we propose to incorporate a novel focus reward in conjuction with the global WGAN reward, to amplify the focus on the distinguishing features of each foreground instance. The focus reward uses guided back propagation maps from an expert task model (\emph{e.g.}~classification) to scale the relative importance of different image regions in the painting process.
Guided backpropagation (GBP) has been shown to be an efficient mechanism for visualizing key image features \cite{springenberg2014striving,nie2018theoretical}. Thus by maximizing the focus reward, we encourage the painting agent to generate canvases with enhanced granularity at key feature locations.

Mathematically, given the normalized guided back-propagation map $\mathcalbf{G}_I \{ \in \{0,1\}^{H \times W}\}$ for the target image, object bounding box coordinates $(x_b,y_b,w_b,h_b)$ and neural alignment model $\mathbf{\Omega}$, we first define the GBP distance $L^\mathcalbf{G}_t$ as,
\begin{align}
    Z^{\mathcalbf{G}_I} = \mathbf{\Omega}(\mathcalbf{G}_I, (x_b,y_b,w_b,h_b)),\\
    L^\mathcalbf{G}_t = \frac{\left \Vert Z^{\mathcalbf{G}_I} \odot \left(Z^I - Z^C_t \right) \right \Vert^2_F}{\left \Vert Z^{\mathcalbf{G}_I} \right \Vert_F},
\end{align}
where $\Vert . \Vert_F$ represents the Frobenius norm. Here we normalize the weighted difference between neurally aligned target and canvas images, using the total number of non-zero pixels in the guided backpropagation map. Thus, the scale of GBP distance $L^\mathcalbf{G}_t$ is invariant to extent of activations in the zoomed key-point importance map $Z^{\mathcalbf{G}_I}$.

The focus reward is then defined as the difference between GBP distances at successive timesteps,
\begin{align}
    r^{focus}_t = L^\mathcalbf{G}_t - L^\mathcalbf{G}_{t+1}.
\end{align}

%------------------------------------------------------------------%
\section{Handling Multiple Foreground Instances}
\label{sec:multiple_foreground}
The semantic guidance pipeline discussed in Section \ref{sec:semantic_guidance}, mainly handles images with a single foreground object instance per image. In this section, we show how the proposed approach can be used to ``learn how to paint'' on datasets depicting multiple foreground objects per image.

At training time, we maintain the bi-level painting procedure described in Section \ref{sec:bilevel}. The action bundle at each timestep describes the brush stroke parameters for the background and one of the foreground instances. The foreground instance for a particular painting episode is kept fixed and is selected with a probability proportional to the total number of pixels covered by that object.
% area of corresponding bounding box

At inference time however, the agent would need to pay attention to all of the foreground instances. Given $N$ total foreground objects, the agent at any timestep $t$ of the painting episode, would choose to predict brush stroke parameters for the foreground class with the highest $l_2$ difference in the corresponding areas on the canvas and the target image. Mathematically, the foreground instance $(u)$ at each timestep $t$ is selected as,
\begin{align}
    u = \argmax_i \Vert \mathcalbf{S}_i \odot (I - C_t) \Vert_F,
\end{align}
where $\mathcalbf{S}_i$ represents the foreground segmentation map for the $i^{th}$ object. We also note that the distinction between foreground and background strokes allows us to perform data augmentation with a specialized dataset to improve the quality of foreground data examples. Thus, in our experiments, we augment the Virtual KITTI dataset with images from Stanford Cars-196 in ratio of 0.8:0.2 while training.
%------------------------------------------------------------------%
% \clear-page 

\section{Experiments}
\label{sec:experiments}

\subsection{Datasets}
\label{sec:datasets}
We use the CUB-200-2011 Birds \cite{WahCUB_200_2011} and  Stanford Cars-196 \cite{KrauseStarkDengFei-Fei_3DRR2013} dataset for performing qualitative evaluation of our method. The above datasets mainly feature one foreground instance per image and hence can be trained using the bi-level semantic guidance pipeline described in Section \ref{sec:semantic_guidance}. We also use the high-fidelity Virtual-KITTI \cite{cabon2020vkitti2} dataset to demonstrate the extension of the proposed method to multiple foreground instances per image.

\textbf{CUB-200-2011 Birds} \cite{WahCUB_200_2011} is a large-scale birds dataset frequently used for benchmarking fine-grain classification models. It consists of 200 bird species with annotations available for class, foreground mask and bounding box of the bird. The dataset features high variation in object background as well as scale, position and the relative saliency of the foreground bird with respect to its immediate surroundings. These properties make it a challenging benchmark for the ``learning to paint'' problem.

\textbf{Stanford Cars-196} \cite{KrauseStarkDengFei-Fei_3DRR2013} is another dataset used for testing fine-grain classification. It consists of 16185 total images depicting cars belonging to 196 distinct categories and having varying 3D orientation. The dataset only provides object category and bounding box annotations. We compute the foreground car masks using the pretrained DeepLabV3-Resnet101 network \cite{chen2017rethinking}.

\textbf{Virtual KITTI} \cite{cabon2020vkitti2} is a high fidelity dataset containing photo-realistic renderings of urban environments from 5 distinct scene backgrounds. Each scene contains images depicting variation in camera location, weather, time of day and density / location of foreground objects. The high variability of these image attributes, makes it a very challenging dataset for training the painting agent. Nevertheless, we demonstrate that our method helps in improving the semantic quality of the generated canvases despite these obstacles.

\subsection{Training Details}
\label{sec:training_details}

\textbf{Neural Renderer.} We closely follow the architecture from Huang \etal \cite{huang2019learning}, while designing the differentiable neural renderer $\mathbf{\Phi}$. Given a batch of random brush stroke parameters $a_t$, the network output $\mathbf{\Phi}[a_t]$ is trained to mimic the rendering of the corresponding Bézier curve on an empty canvas. The training labels are generated using an automated graphics module and the renderer is trained for $4 \times 10^5$ iterations with a batch size of 64.

\textbf{Learning foreground mask and bounding box.} A key component of the semantic guidance pipeline is foreground segmentation and bounding box prediction. We use a fully convolutional network, with separate heads to predict a per-pixel foreground probability map and the coordinates of the bounding box.
The foreground mask prediction is trained with the standard cross-entropy loss $L_{fg}$, while the bounding box coordinates are learned using Smooth L1 \cite{girshick2015fast} regression loss $L_{bbox}$.

\textbf{Expert model for Guided Backpropagation.} 
We use the pretrained fine-grain classification NTS-Net model \cite{yang2018learning} as the expert network used for generating guided backpropagation maps on the CUB-200-2011 birds dataset. Note that we use NTS-Net due the easy accessibility of the pretrained model. We expect that using a more state of the art model like \cite{ge2019weakly} would lead to better results with the focus reward.

The expert model for the Standford Cars-196 dataset is trained in conjunction with the reinforcement learning agent, with an EfficientNet-B0 \cite{tan2019efficientnet} backbone network. The EfficientNet architecture allows us to limit the total number of network parameters while respecting the memory constraints for a NVIDIA GTX 2080 Ti. The expert model is trained for a total of 200 epochs with a batch size of 64. EfficientNet-B7 model pretrained on ImageNet \cite{deng2009imagenet} dataset, is used as the expert for the Virtual KITTI dataset.

\textbf{Overall Training.} The reinforcement learning agent follows an actor-critic architecture. The actor predicts the policy function $\pi(a|s)$, while the critic computes the value function $V(s)$. The agent is trained using model-based DDPG \cite{lillicrap2015continuous} with the following policy and value loss,
\begin{align}
    &L_{actor} = -\mathbf{E}_{s_t,a_t}\left[ r(s_t,a_t) + \gamma V(s_{t+1}) \right], \\
    &L_{critic} = \mathbf{E}_{s_t,a_t} \left[ (r(s_t,a_t) + \gamma V(s_{t+1}) - V(s_{t}) )^2 \right]
\end{align}
where $\gamma$ is the discount factor and the final reward function $r(s_t,a_t)$ is computed as the weighted sum of the foreground, background and focus rewards,
\begin{align}
    r(s_t,a_t) = r^b_t + \eta \ r^f_t + \nu r^{focus}_t,
\end{align}
where $\eta,\nu$ are hyperparameters. A hyper-parameter selection of $\{\eta=2,\nu=10\}$ was seen to give competitive results for our experiments. The model-based RL agent is trained for a total of 2M iterations with a batch size of 96.

%------------------------------------------------------------------%
% \clearpage
\subsection{Results}
\label{sec:results}
\begin{figure*}[ht]
% \vskip 0.2in
\begin{center}
\centerline{\includegraphics[width=0.95\linewidth]{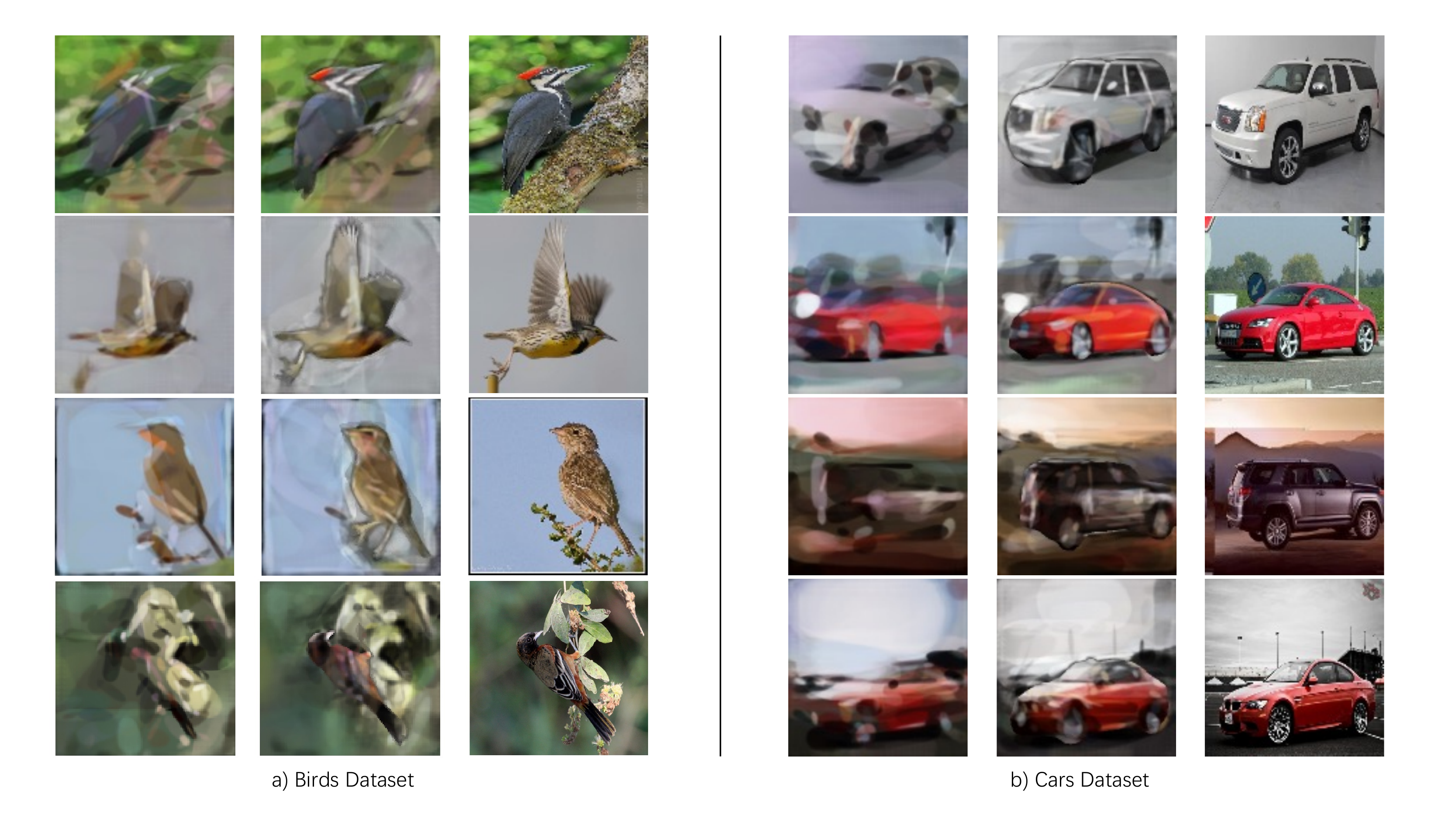}}
\caption{\textbf{Results on CUB-200 Birds and Stanford-Cars196 Datasets.} Left: Huang \etal \cite{huang2019learning}, Middle: Canvas generated using Semantic Guidance pipeline (Ours), Right: the original target image. We clearly see that our method results in enhanced foreground saliency and achieves better granularity of key object features.}
\label{fig:bird_results}
\end{center}
\vskip -0.3in
\end{figure*}

\begin{figure}[ht]
\vskip 0.1in
\begin{center}
% \centerline{\includegraphics[width=0.98\linewidth,height=4cm]{bird_results1.pdf}}
\centerline{\includegraphics[width=\linewidth]{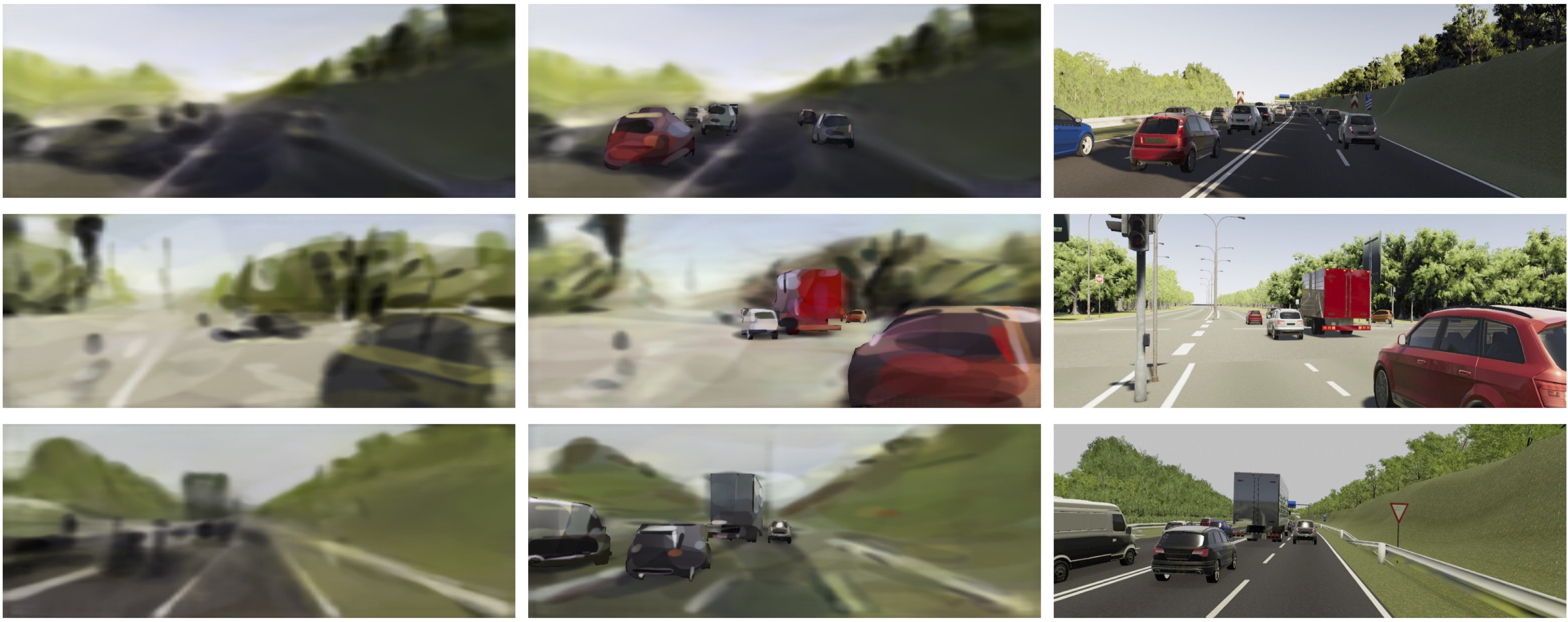}}
\caption{\textbf{Results on Virtual KITTI.} Left: Baseline \cite{huang2019learning}, Middle: Canvas generated using Semantic Guidance pipeline (Ours), Right: target image. By developing a distinction between foreground and background strokes, our method better captures the color / saliency of visually small foreground vehicles.}
\label{fig:vkitti_results}
\end{center}
\vskip -0.2in
\end{figure}

We compare our method with the baseline ``learning to paint'' pipeline from Huang \etal \cite{huang2019learning} which uses an action bundle containing 5 consecutive brush strokes. In order to provide a fair comparison, we use the same overall bundle size but divide it among foreground and background strokes in the ratio of 3:2. That is, the agent at each timestep predicts 3 foreground and 2 background brush strokes.

\textbf{Improved foreground saliency.} Fig.~\ref{fig:bird_results} shows the results\footnote{Please refer supp. material for further quantitative results.} for the CUB-200 Birds and Stanford-Cars196 dataset. We clearly see that our method leads to increased saliency of foreground objects, especially when the target object is partly camouflaged by its immediate surroundings (refer Fig.~\ref{fig:bird_results}a, row-4 and Fig.~\ref{fig:bird_results}b, row-3). This increased contrast between foreground and background perception, results directly from our semantically guided bi-level painting process and the neural alignment model. 

\textbf{Enhanced feature granularity.} We also observe that canvases generated using our method show improved focus on key object features as compared to the baseline. For instance, the red head-feather, which is an important feature of pileated woodpecker (refer Fig.~\ref{fig:bird_results}a: row-1), is practically ignored by the baseline agent due to its small size. The proposed guided backpropagation based focus reward, helps in amplifying the importance of this key feature in the overall reward function. Similarly, our method also leads to improved depiction of wing patterns and claws in (Fig.~\ref{fig:bird_results}a: row-2), the small eye region, feather marks in (Fig.~\ref{fig:bird_results}a: row-3) and car headlights, wheel patterns in (Fig.~\ref{fig:bird_results}b: row-1,2).

\textbf{Multiple foreground instances.} We use the Virtual-KITTI dataset and the extended training procedure outlined in Section \ref{sec:multiple_foreground}, to demonstrate the applicability of our method on images with multiple foreground instances. Note that due to computational limits and the nature of ground-truth data, we stick to vehicular foreground classes like cars, vans, buses \emph{etc}, for our experiments. Results are shown in Fig.~\ref{fig:vkitti_results}. We observe that due to the dominant nature of image backgrounds in this dataset, the baseline agent fails to accurately capture the presence / color spectrum of the foreground vehicles. In contrast, our bi-level painting procedure learns a distinction between foreground and background strokes in the training process itself, and thus provides a much better balance between foreground and background depiction for the target image.

%------------------------------------------------------------------%
\section{Analysis}
\label{sec:analysis}

\subsection{Ablation Study: Isolating Impact of Focus Loss}
\label{sec:ablation_focus}

In this section, we design a control experiment in order to isolate the impact of focus reward proposed in Section \ref{sec:focus_reward}. To this end, we construct a modified birds dataset from CUB-200-2011 dataset. We do this by first setting the background image pixels to zero, which alleviates the need for the bi-level painting procedure. We next eliminate the need for the neural alignment model by cropping the bounding box for each bird. The resulting dataset is then used to train the baseline \cite{huang2019learning}, and a modified semantic guidance pipeline trained only using a weighted combination of the WGAN reward \cite{huang2019learning} and the focus reward $r^{focus}_t$,
\begin{align}
    r(s_t,a_t) = r^{wgan}_t + \kappa \ r^{focus}_t,
\end{align}
where $\kappa=0$ represents baseline model without the focus loss. We then analyse the effect on the resulting canvas as the weightage $\kappa$ of the focus reward is increased.
All models are trained for 1M iterations with a batch size of 96.

\begin{figure}[hb!]
\vskip -0.2in
\begin{center}
% \centerline{\includegraphics[width=0.98\linewidth,height=4cm]{bird_results1.pdf}}
\centerline{\includegraphics[width=0.98\linewidth]{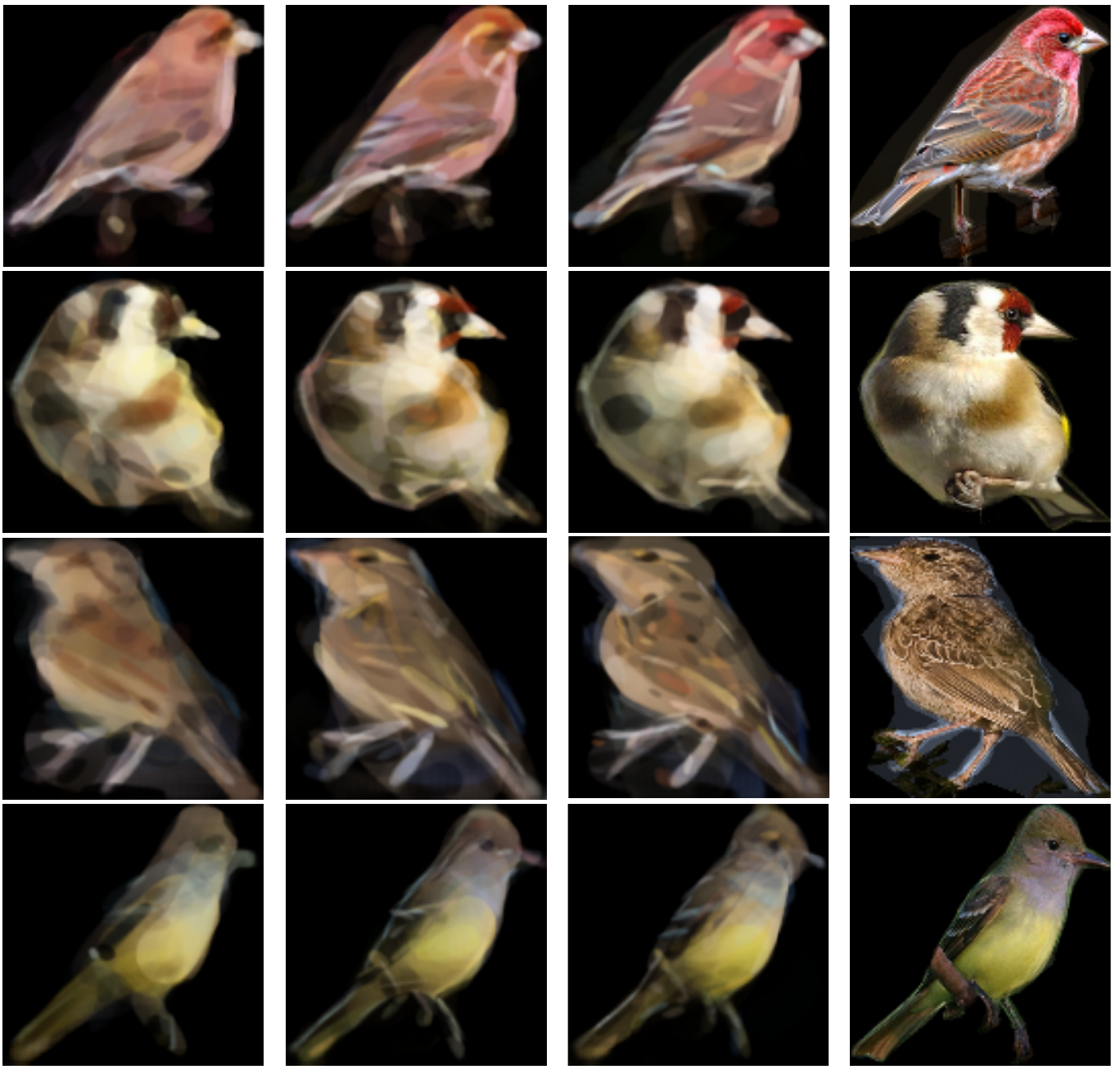}}
\caption{\textbf{Ablation results for focus reward.} (Column 1-3): From left to right, the painted canvases for $\kappa = 0,5,10$ respectively, where $\kappa=0$ represents the baseline \cite{huang2019learning}. (Column-4): the target image from modified birds dataset (refer Sec. \ref{sec:ablation_focus}). We see a clear increase in the amount of finer feature details like wing texture, density of eyes \emph{etc}, as the weightage of focus loss is increased.}
\label{fig:gbp_results}
\end{center}
\vskip -0.4in
\end{figure}

Fig.~\ref{fig:gbp_results} describes the modified training results. We clearly see that while the baseline \cite{huang2019learning} trained with wgan reward captures the overall bird shape and color, it fails to accurately pay attention to finer bird features like texture of the wings (row 1,3,4), density of eyes (row 3,4) and sharp color contrast (red regions near the face for row 1,2). 
We also observe that the granularity of the above discussed features in the painted canvas, improves as the weightage $\kappa$ of the focus reward is increased.

\subsection{Analysing Effect of Semantic Guidance on Painting Sequence}
\label{sec:paint_seq}

\begin{figure*}[ht]
% \vskip 0.2in
\begin{center}
% \centerline{\includegraphics[width=0.98\linewidth,height=4cm]{bird_results1.pdf}}
\centerline{\includegraphics[width=0.87\linewidth]{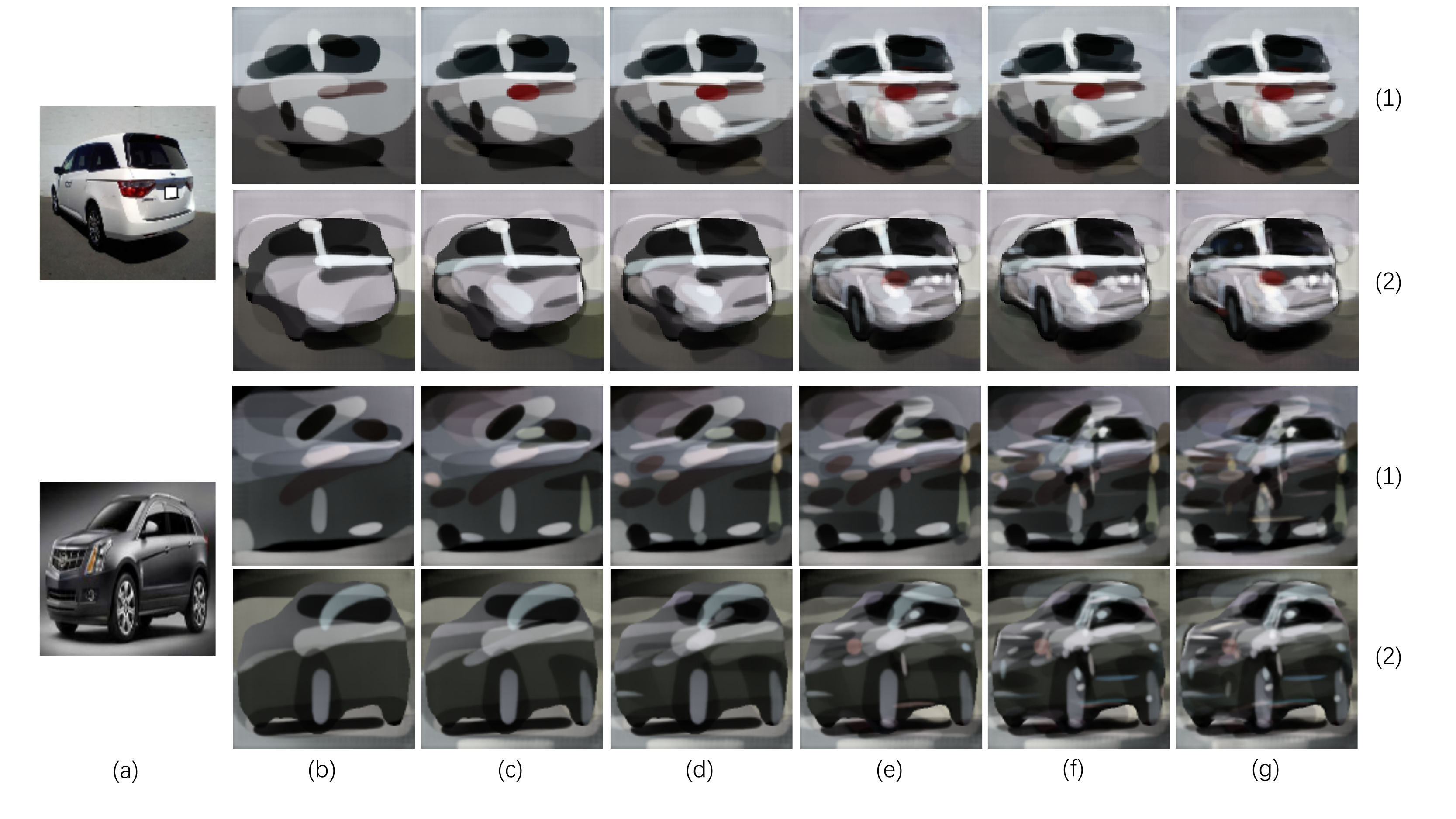}}
\caption{\textbf{Effect of Semantic Guidance on Painting Sequence.} (1) Baseline \cite{huang2019learning}, (2) Semantic Guidance (Ours). For each target image in (a), (b-g) represent the canvas state after 10, 20, 30, 50, 100, 200 brush strokes respectively. We observe that there is huge difference between the painting styles of the two agents. In contrast to the baseline agent (which follows a bottom-up approach), the top-down painting style of our method offers better resemblance with a human painter.}
\label{fig:paint_seq}
\end{center}
\vskip -0.3in
\end{figure*}

Recall that the main goal of the ``learning to paint'' problem, is to make the machine paint in a \emph{manner similar to a human painter}. Thus, the performance of a painting agent should be measured, not only by the resemblance between the final canvas and the target image, but also by the similarity of the corresponding painting sequence with that of a human painter. In this section, we demonstrate that unlike previous methods, semantic guidance helps the reinforcement learning agent adopt a painting trajectory that is highly similar to the human painting process.

In order to do a fair comparison of agent trajectories between our method and the baseline \cite{huang2019learning}, we select test images from the Stanford Cars-196 dataset, such that the final canvases from both methods are equally similar to the target image. That is, the $l_2$\footnote{We note that, in general $l_2$ distance may not be a reliable measure of semantic similarity between two images. As shown in Fig.~\ref{fig:paint_seq}, two canvases can be qualitatively quite different while having similar $l_2$ distance with the target image.} distance between the final canvas and the target image is similar for both methods.

Results are shown in Fig.~\ref{fig:paint_seq}. We can immediately observe a stark difference between the painting styles of the two agents. The standard agent
displays bottom-up image understanding, and proceeds to first paint visually distinct car edges / parts like windows, red tail light, black region near the bottom of the car \emph{etc}. In contrast, the semantically guided agent follows a more human-like top-down approach, wherein it first  begins with a rough structural outline for the car and only then focuses on other structurally non-relevant parts. For instance, in the first example from Fig.~\ref{fig:paint_seq}, the semantic guidance agent adds color to the tail-light only after finishing painting the overall structure of the car. On the other hand, the red brush stroke for the tail-light region is painted quite early by the baseline agent, even before the overall car structure begins to emerge on the canvas. 

We thus note that the semantically guided agent resembles the human painting style on two broad levels. \emph{1) On the canvas level}, the bi-level procedure allows the painting agent to learn different stroke patterns for semantically distinct image regions (as is done by humans). 
% This is in contrast to [\citeg{15}] where the RL agent directly operates on the low-level features. 
\emph{2) On the object level}, expert guidance and specialization of the foreground strokes to a specific object class (\emph{e.g.}~cars) leads to a top-down painting sequence. That is, the model first pays attention to high-level structural features which are shared by several instances of the foreground object class, and only then focuses on finer instance-specific lower-level features.

%------------------------------------------------------------------%
\section{Conclusion}
\label{sec:conclusion}

In this paper, we propose a semantic guidance pipeline for the ``learning to paint'' problem. Our method incorporates semantic segmentation to propose a bi-level painting process, which helps in learning a distinction between foreground and background brush stroke patterns. 
% This distinction is seen to lead to increased saliency of foreground objects. 
We also introduce a guided backpropagation based focus reward, to increase the granularity and importance of small but distinguishing object features in the final canvas. The resulting agent successfully handles variations in position, scale and saliency of foreground objects, and develops a top-down painting style which closely resembles a human painter. 

\textbf{Acknowledgements.} This work was supported by ARC Discovery Early Career Researcher Award (DE200101283) and ARC Discovery Project (DP210102801).

%------------------------------------------------------------------%
% references
\clearpage
{\small
\bibliographystyle{ieee_fullname}
\bibliography{egbib}
}

%------------------------------------------------------------------%
% \clearpage
\appendix
\appendixpage
%------------------------------------------------------------------%
\section{Quantitive Results}
\label{sec:quantitative_results}

\subsection{Measuring Semantic Similarity.}
\label{sec:semantic_similarity}

\begin{figure*}[ht!]
% \vskip 0.2in
\begin{center}
\centerline{\includegraphics[width=0.95\linewidth]{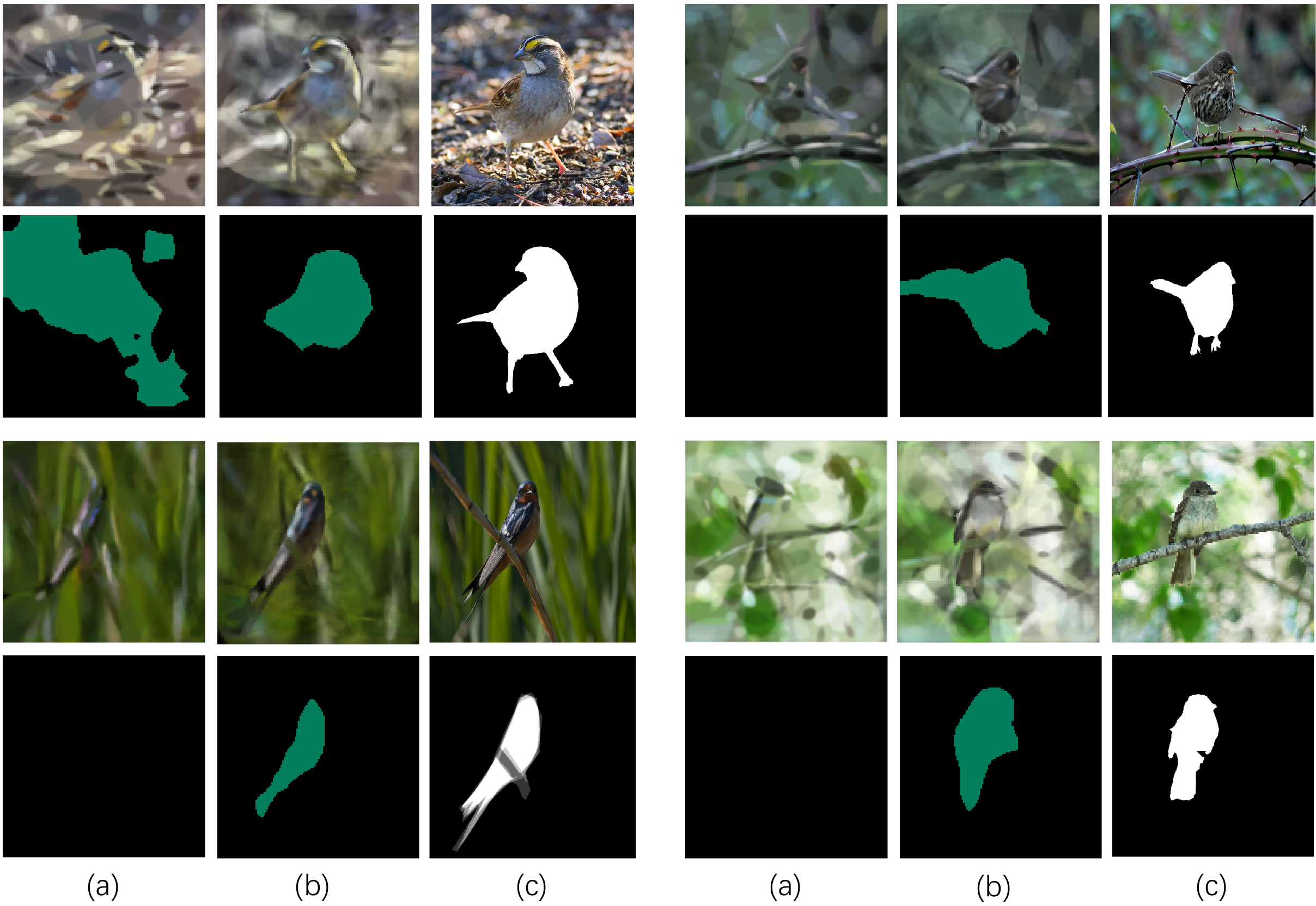}}
\caption{\textbf{Analysing Semantic Similarity}.(a) Huang \etal \cite{huang2019learning}, (b) Semantic Guidance (Ours), (c) the target image. The bottom row for each example represents the semantic segmentation maps for the images shown in the top row. We clearly see that the canvases painted using our method generate semantic segmentation maps which are much closer to the ground truth foreground segmentation masks. We also note that, for target images with low foreground background contrast, the segmentation maps for baseline canvases (a) fail to even indicate the presence of the foreground object.}
\label{fig:semantic_results}
\end{center}
% \vskip -0.1in
\end{figure*}

\begin{table}[H]
\begin{center}
\begin{tabular}{|c|c|c|}
\hline
Method & Accuracy & IoU \\
\hline\hline
Huang \etal \cite{huang2019learning} & 45.41 & 27.21 \\
Semantic Guidance (Ours)  & \textbf{69.26} & \textbf{48.15}\\
\hline
\end{tabular}
\end{center}
\caption{\textbf{Semantic Similarity Results on CUB-200 Birds.} The semantic segmentation maps (refer Appendix \ref{sec:semantic_similarity}) for the canvases generated using our method, result in much better segmentation accuracy and Intersection over Union (IoU) scores.}
\label{tab:semantic_results}
\end{table}

% The inadequacy of pixel-wise $l2$ distance in capturing semantic similarity poses a major challenge in performing a qualitative evaluation of our method.

The inadequacy of the frequently used pixel-wise $l2$ distance \cite{ganin2018synthesizing,huang2019learning} in capturing semantic similarity, poses a major challenge in performing a quantitative evaluation of our method.
In order to address this, we present a novel approach to quantitatively evaluate the semantic similarity between the generated canvases and the target image. To this end, we use a pretrained DeeplabV3-ResNet101 model \cite{chen2017rethinking} to compute the semantic segmentation maps for the final painted canvases for both Huang \etal \cite{huang2019learning} and the Semantic Guidance (Ours) approach. The detected segmentation maps for both methods are then compared with the ground truth foreground masks for the target image. 

Results are shown in Fig.~\ref{fig:semantic_results}. We clearly see that our method learns to paint canvases with semantic segmentation maps having high resemblance with the ground truth foreground masks for the target image. In contrast, the canvases generated using the baseline \cite{huang2019learning} show low foreground saliency. This sometimes results in the pretrained segmentation model \cite{chen2017rethinking} even failing to detect the presence of the foreground object. Note that the semantic guidance pipeline does not directly train the RL agent to mimic the segmentation maps of the original image.

We also provide a more quantitative evaluation of the quality of detected semantic segmentation maps for both methods in Table \ref{tab:semantic_results}. The accuracy scores are reported on the test set images and represent the percentage of foreground pixels which are correctly detected in the segmentation map of a given canvas. We observe that our method leads to huge improvements in the semantic segmentation accuracy and IoU values for the painted canvases.

The above qualitative and quantitative results conclusively demonstrate that the semantic guidance pipeline leads to huge gains ($\sim$ 25\%) in preserving the underlying semantics of a given scene. 

% The inadequacy of pixel-wise $l2$ distance in capturing semantic similarity poses a direct hindrance in performing a qualitative evaluation of our method. To this end, we 

\subsection{Enhanced Foreground Resemblance}
\label{sec:foreground_resemblance}

\begin{table}[h!]
\begin{center}
\begin{tabular}{|c|c|c|}
\hline
Method & Foreground L2 Distance\\
\hline\hline
Huang \etal \cite{huang2019learning} &  8.43 \\
Semantic Guidance (Ours)  & \textbf{7.81} \\
% \hline \hline\hline
% Method & Overall L2 Distance\\
% \hline\hline
% Huang \etal \cite{huang2019learning} &  \textbf{17.96} \\
% Semantic Guidance (Ours)  & 18.02 \\
\hline
\end{tabular}
\end{center}
\caption{\textbf{Foreground Resemblance Results on CUB-200 Birds.} Our approach leads to a lower average L2 distance between the foreground regions of the target image and the generated canvas.}
% We note that while the overall L2 distance for both methods is similar, 
\label{tab:semantic_results}
\end{table}

%------------------------------------------------------------------%
\section{Implementation of Neural Alignment Model}
The  neural alignment model is implemented by replacing the localization net of a standard spatial transformer network \cite{jaderberg2015spatial} with the bounding box prediction network. We also note that the $3 \times 2$ affine matrix defined in Eq.~11 of the main paper, represents the ideal affine mapping operation from input to output image coordinates. However, the affine matrix used for practical implementations may vary based on the conventions of the used deep learning framework. For our implementation (in pytorch), we compute the affine matrix for the spatial transformer network as follows,
\begin{align}
    \Tilde{A} = 
    \begin{bmatrix}
    \Tilde{w}_b & 0 & 2\Tilde{x}_b + \Tilde{w}_b - 1\\
    0 & \Tilde{h}_b& 2\Tilde{y}_b + \Tilde{h}_b - 1
    \end{bmatrix}^T ,
\end{align}

where $(\Tilde{x}_b,\Tilde{y}_b ,\Tilde{w}_b ,\Tilde{h}_b )$ are the normalized bounding box coordinates of the foreground object.
% \begin{align}
%     A = 
%     \begin{bmatrix}
%     w_b/W & 0 & (2x_b + w_b)/W - 1\\
%     0 & h_b/H & (2y_b + h_b)/H - 1
%     \end{bmatrix}^T ,
% \end{align}
% where $(x_b,y_b,w_b,h_b)$ are the bounding box coordinates of the foreground object and $(H,W)=(128,128)$ represent the overall image dimensions. 

%------------------------------------------------------------------%
\section{Note on Over-painting Phenomenon}
We note that while the proposed semantic guidance pipeline results in huge improvements in enhancing foreground object saliency and increasing the granularity of key features in the painted image, we do observe minor background artifacts for images with plain backgrounds. This occurs because as part of the bilevel painting procedure, both foreground and background brush strokes are working simultaneously in an action bundle. Thus for images with high contrast in complexities of foreground and
background, the background strokes are forced to slightly \emph{overpaint}
while the foreground strokes draw the in-focus object. This \emph{overpainting} phenomenon was seen to cause minor artifacts in plain image backgrounds as can be seen in (Fig.~\ref{fig:bird_results}a; row-3) of the main paper. The above mentioned artifacts can be reduced by adaptively balancing the number of foreground / background strokes in an action bundle, based on the WGAN distances for the foreground and background image regions. However, the same is out of scope of the current paper and we thus leave it here as a possible future research directive.

\end{document}